\documentclass[runningheads]{llncs}
\pdfoutput=1
\usepackage{graphicx}
\usepackage{amsmath,amssymb} 
\usepackage{color}
\usepackage{cite}
\usepackage{subfig}
\usepackage[pagebackref=true,breaklinks=true,letterpaper=true,colorlinks,bookmarks=false]{hyperref}
\usepackage{booktabs}
\usepackage{multirow}
\usepackage{makecell}
\usepackage{amsmath, bm}
\usepackage{epstopdf}
\usepackage{epsfig}
\newcommand{\tabincell}[2]{\begin{tabular}{@{}#1@{}}#2\end{tabular}}
\begin{document}
	
	\title{Appearance-Based Gaze Estimation Using Dilated-Convolutions\thanks{Supported by the Innovation and Technology Fund of Hong Kong under grant ITS/406/16FP.}} 
	\titlerunning{Dilated-Convolutions for Gaze Estimation} 
	
	%
	\author{Zhaokang Chen\orcidID{0000-0003-0237-0358} \and
		Bertram E. Shi\orcidID{0000-0001-9167-7495}}
	

	\authorrunning{Z. Chen and B. E. Shi} 
	
	
	\institute{The Hong Kong University of Science and Technology, Hong Kong SAR\\
		\email{\{zchenbc,eebert\}@ust.hk}}
	
	\maketitle
	
	\begin{abstract}
		Appearance-based gaze estimation has attracted more and more attention because of its wide range of applications. The use of deep convolutional neural networks has improved the accuracy significantly. In order to improve the estimation accuracy further, we focus on extracting better features from eye images. Relatively large changes in gaze angles may result in relatively small changes in eye appearance. We argue that current architectures for gaze estimation may not be able to capture such small changes, as they apply multiple pooling layers or other downsampling layers so that the spatial resolution of the high-level layers is reduced significantly. To evaluate whether the use of features extracted at high resolution can benefit gaze estimation, we adopt dilated-convolutions to extract high-level features without reducing spatial resolution. In cross-subject experiments on the Columbia Gaze dataset for eye contact detection and the MPIIGaze dataset for 3D gaze vector regression, the resulting Dilated-Nets achieve significant (up to 20.8\%) gains when compared to similar networks without dilated-convolutions. Our proposed Dilated-Net achieves state-of-the-art results on both the Columbia Gaze and the MPIIGaze datasets.
		
		\keywords{Appearance-based gaze estimation  \and Dilated-convolutions.}
	\end{abstract}
	
	\section{Introduction}
	Gaze tracking has long been considered as an important research topic, as it has various promising real-world applications, such as gaze-based interfaces~\cite{pi2017probabilistic,chen2018using}, foveated rendering in virtual reality~\cite{patney2016towards}, behavioral analysis~\cite{hoppe2018eye} and human-robot interaction~\cite{huang2016anticipatory}. Early gaze tracking techniques required strong constraints, e.g., facing the tracker frontally and keeping the positions of eyes inside a certain region. These constraints limited the applications to relatively controlled environments. In order to apply gaze tracking in real-world and more flexible environments, researchers proposed many novel methods to alleviate these constraints and move towards unconstrained gaze tracking, e.g., ~\cite{schneider2014manifold,sugano2014learning,zhang2017mpiigaze,baltrusaitis2018openface,ranjan2018light,deng2017monocular,wang2018slam}.

	Unconstrained gaze tracking refers to calibration-free, subject-, viewpoint- and illumination-independent gaze tracking~\cite{ranjan2018light}. Appearance-based gaze estimation is a promising approach to unconstrained gaze tracking. It estimates the 2D gaze target position on a given plane or 3D gaze angles based on the images captured by RGB sensors. The key advantage of this method is that it does not require expensive custom hardware but off-the-shelf cameras, which are inexpensive and commonly available nowadays. However, it is a very challenging problem as it needs to address several factors, such as differences in individual appearances, head-eye relationships, gaze ranges and illumination conditions~\cite{zhang2017mpiigaze}.
	
	In recent years, with the success of deep convolutional neural networks (CNNs) in the field of computer vision, researchers have started to apply deep CNNs to appearance-based gaze tracking. Thanks to the large number of publicly-available high quality real and synthetic datasets~\cite{zhang2015appearance,krafka2016eye,sugano2014learning,smith2013gaze,funes2014eyediap,shrivastava2017learning}, deep CNNs have demonstrated good performance, but there is still room for improvement.
	
	In this article, we propose to improve the accuracy of appearance-based gaze estimation by extracting higher resolution features from the eye images using deep neural networks. Given the fact that eye images with different gaze angles may differ only by a few pixels (see Fig.\ref{fig:1}), we argue that extracting features at high resolution could improve accuracy by capturing small appearance changes. To extract high-level features at high spatial resolution, we applied dilated-convolutions (alternatively, atrous-convolutions), which efficiently increase the receptive field sizes of the convolutional filters without reducing the spatial resolution. The main contributions of this article are that we propose Dilated-Net and quantitatively evaluate the use of high resolution features on the Columbia Gaze~\cite{smith2013gaze} and MPIIGaze datasets~\cite{zhang2015appearance} through Dilated-Net. In cross-subject experiments, the proposed Dilated-Net outperform CNNs with similar architecture significantly from $3.2\%$ to $20.8\%$ depending on the task. It achieves state-of-the-art results on both datasets. The results demonstrate that the use of high resolution features benefit gaze estimation.
	
	\section{Related Work}
	\subsection{Appearance-Based Gaze Estimation}
	Methods for appearance-based gaze estimation learn a mapping function from images to gaze estimate, where the estimation target is normally defined as either a gaze target in a give plane (2D estimation) or a gaze direction vector in camera coordinates (3D estimation). Appearance-based methods are attracting more and more attention as they use inputs from off-the-shelf cameras, which are widely available. Given enough training data, they may be able to achieve unconstrained gaze estimation.
	
	Several methods in computer vision have been applied to this problem, e.g., Random Forests~\cite{sugano2014learning}, k-Nearest Neighbors~\cite{sugano2014learning,schneider2014manifold}, Support Vector Regression~\cite{schneider2014manifold} and, recently, deep CNNs. Zhang \textit{et al}. proposed the first deep CNN to estimate 3D gaze angles~\cite{zhang2015appearance,zhang2017mpiigaze}. Their network takes the left eye image and the estimated head pose angles as input. They showed that the use of deep CNNs trained on a large amount of data improved accuracy significantly. To employ information outside the eye region, Krafka \textit{et al}. proposed a multi-region CNN that takes an image of the face, images of both eyes and a face grid as input to estimate the gaze target on phone and tablet screens~\cite{krafka2016eye}. Zhang \textit{et al}. proposed a network that takes the full face image as input and uses a spatial weights method to emphasize features extracted from particular regions~\cite{zhang2017s}. This work has shown that regions of the face other than the eyes also contain information about gaze angles. To further improve the accuracy, other work has concentrated on estimating the head-eye relationships. Ranjan \textit{et al}. applied a branching architecture, where parameters are switched by clustering head pose angles into different groups~\cite{ranjan2018light}. Deng and Zhu trained two networks to estimate the head pose angles in camera coordinates and gaze angles in head coordinates separately. The final gaze angles in camera coordinate were obtained by combing the estimates geometrically~\cite{deng2017monocular}.
	
	Instead of estimating the continuous gaze directions, some work considered gaze tracking as a classification problem by dividing the gaze directions into certain blocks. For example, George and Routray applied a CNN to classify eye images into 3 or 7 target regions~\cite{george2016real}. The binary classification problem, referred as gaze locking or eye contact detection, detects whether the user is looking at the camera. A Support Vector Machine (SVM)~\cite{smith2013gaze}, Random Forests~\cite{ye2015detecting} and a CNN with multi-region input~\cite{parekh2017eye} have been applied to this problem. 
	
	While the recent trend has been to investigate how information from regions other than the eyes can benefit gaze estimation~\cite{zhang2017s,ranjan2018light,deng2017monocular}, 
	here we focus on how better features extracted from the eye images can be used to benefit multi-region networks.
	
	\begin{figure}[!t]
		\centering
		\subfloat[$(10^\circ \rm H, 0^\circ \rm V)$]{		
			\includegraphics[width=3.6cm]{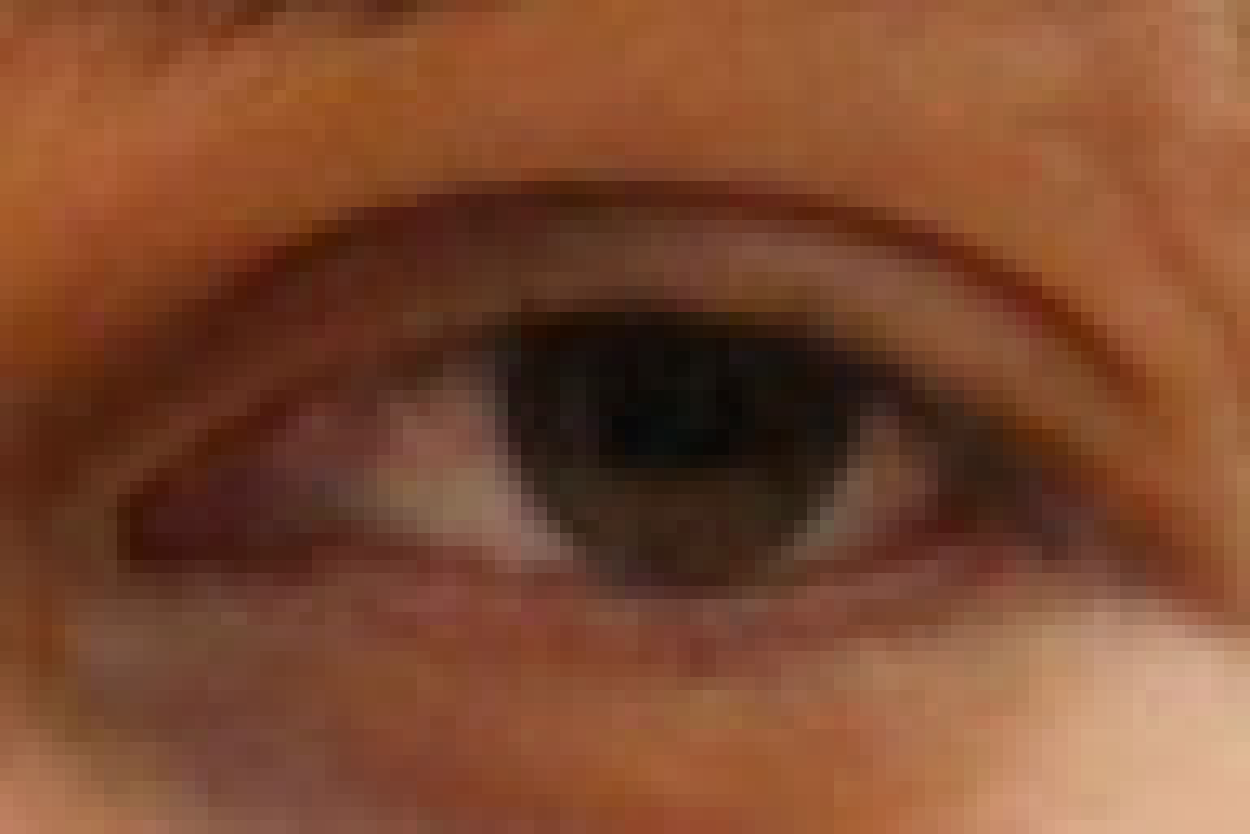}
			\label{fig:1-a}	
		}
		\subfloat[$(15^\circ \rm H, 0^\circ \rm V)$]{		
			\includegraphics[width=3.6cm]{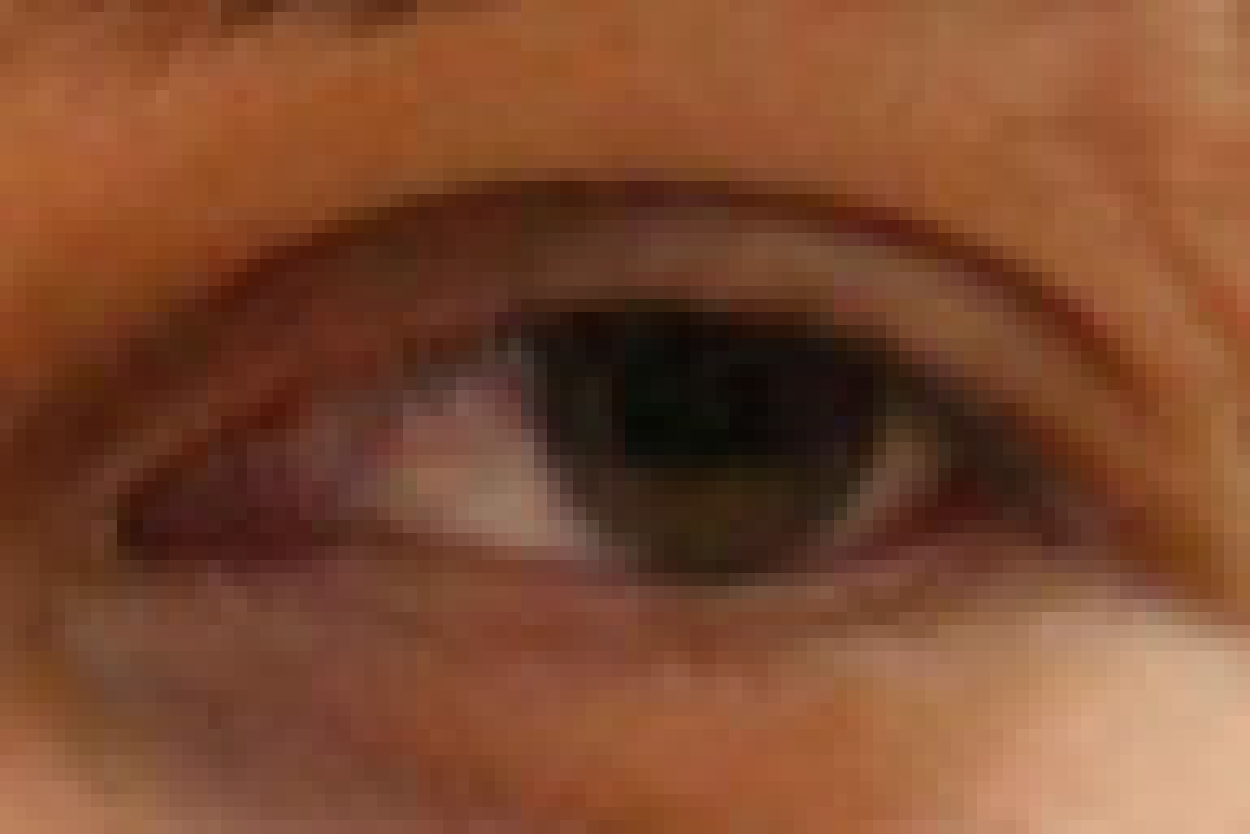}
			\label{fig:1-b}
		}
		\subfloat[Absolute differences]{
			\includegraphics[width=3.6cm]{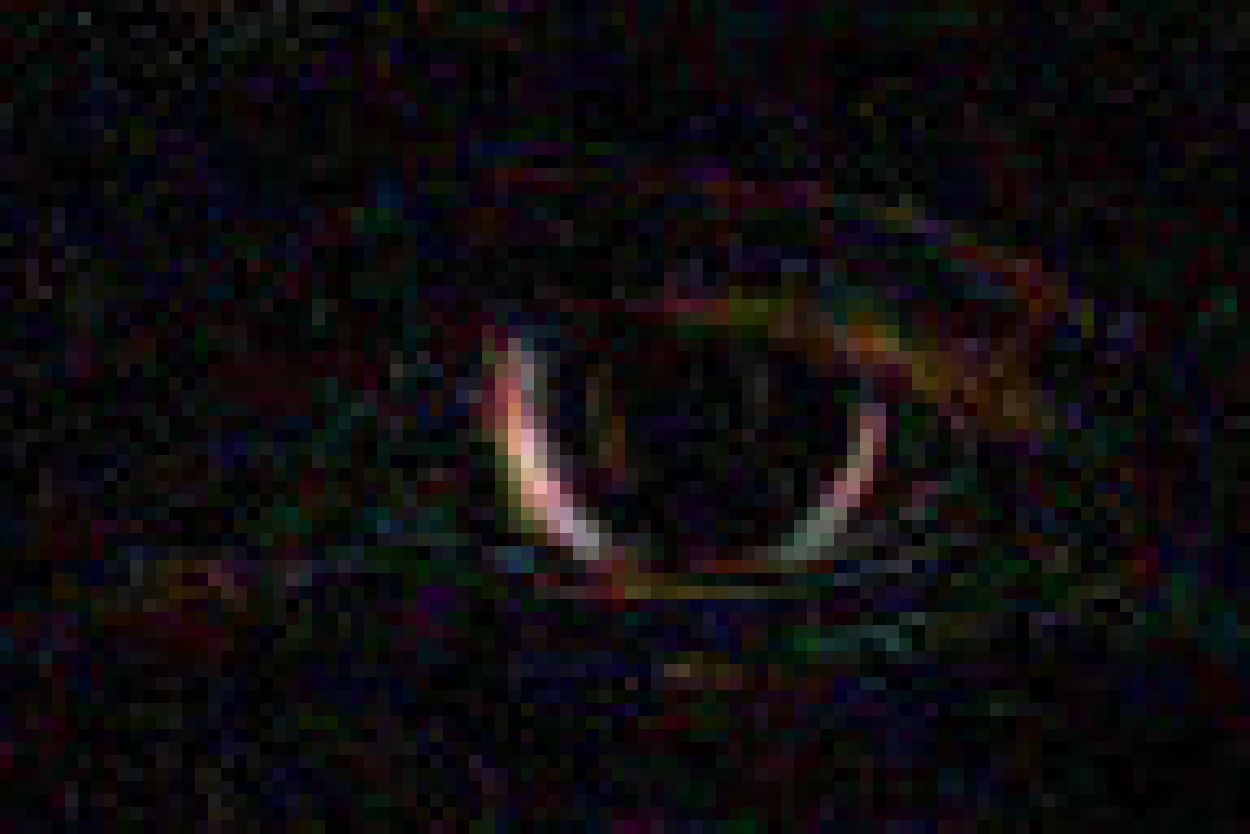}
			\label{fig:1-c}
		}
		\caption{Images of two left eyes and their difference from the Columbia Gaze dataset~\cite{smith2013gaze}. (a) Left eye image with $10^\circ$ horizontal and $0^\circ$ vertical gaze angle. (b) Left eye image with $15^\circ$ horizontal and $0^\circ$ vertical gaze angle. (c) The absolute difference between (a) and (b) (scaled for better illustration).}
		\label{fig:1}
	\end{figure}
	
	\subsection{Dilated-Convolutions}
	Dilated-convolutions were first introduced in the field of computer vision to extract dense features for dense label prediction, i.e., semantic segmentation~\cite{yu2015multi,chen2014semantic}. Given a convolutional kernel of size $N\times M\times K$ (\textit{height}$\times$\textit{width}$\times$\textit{channel}), the key idea of dilated-convolutions is to insert spaces (zeros) between the weights so that the kernel covers a region larger than $N\times M$. Therefore, dilated-convolutions increase the size of receptive field without reducing the spatial resolution nor increasing the number of parameters. Comprehensive studies of dilated-convolution in semantic segmentation were reported in~\cite{wang2017understanding,chen2018deeplab}, where the results show that dilated-convolutions improve performance significantly. Recently, Yu \textit{et al}. proposed dilated residual networks~\cite{yu2017dilated} and showed that they outperform their non-dilated counterparts in image classification and object localization on the ImageNet dataset~\cite{deng2009imagenet}.
	
	\section{Methodology}
	
	\subsection{Issue of Spatial Resolution}
	When a person looks at two different locations with his/her head fixed, the appearance of the eyes changes. However, these differences can be subtle, as shown in Fig.~\ref{fig:1}. A $5^\circ$ horizontal difference only results in differences at a few pixels. Other small changes, e.g., in the openness and the shape of the eyes, also contain information about gaze direction. Intuitively, extracting high-level features at high resolution will better capture these subtle differences.
	
\begin{figure}[!t]
		\centering
		\subfloat[$3\times3$ convolution $+~2\times2$ max-pooling $+~3\times3$ convolution]{		
			\includegraphics[width=4.5cm]{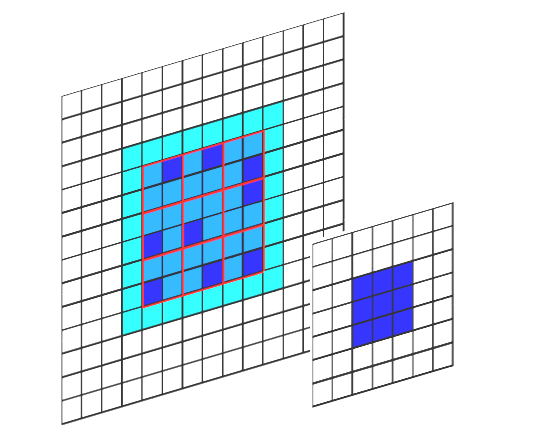}
			\label{fig:2-a}	
		}
		\\
		\subfloat[$3\times3$ dilated-convolution $(1, 1)$
		$+~3\times3$ dilated-convolution $(2, 2)$
		]{		
			\includegraphics[width=4.5cm]{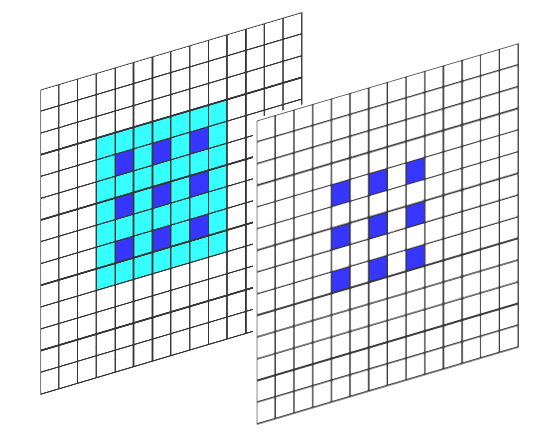}
			\label{fig:2-b}
		}
		\qquad
		\subfloat[$3\times3$ convolution $+~3\times3$ convolution]{
			\includegraphics[width=4.5cm]{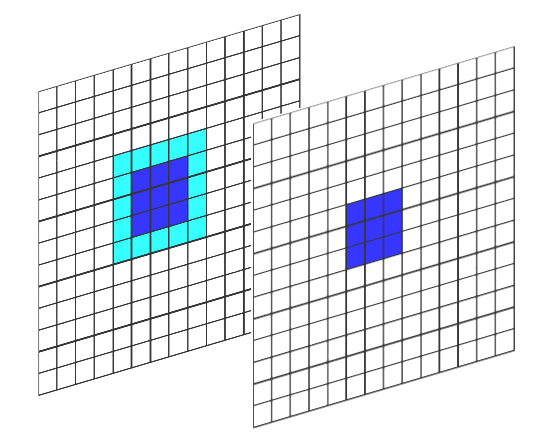}
			\label{fig:2-c}
		}
		\caption{Receptive fields for three different combinations of layers. The grid on the left represents the lower-level feature map. The grid on the right represents the output of the max-pooling (a) or the convolution (b,c). Locations in dark blue show locations weighted by the convolution operating on the right grid, and the corresponding locations in the left grid. Light blue shows the effective size of the RF in the lower layer due to the first convolution. The strides for convolutions and dilated-convolutions are 1 and the stride for max-pooling are 2.}
		\label{fig:2}
	\end{figure}

	Most current CNN architectures use multiple downsampling layers, e.g., convolutional layer with large stride and pooling layers. In this article, we use max-pooling layers as an example for discussion because they are commonly used both in general and in gaze estimation~\cite{zhang2017mpiigaze,zhang2017s,krafka2016eye,deng2017monocular,ranjan2018light}. Similar considerations apply for convolutional layers with large stride. The use of max-pooling layers progressively reduces the spatial resolution of feature maps. This enables the networks to tolerate small variations in position, increases the effective size of receptive field (RF) at higher layers and reduces the number of parameters in the networks.  However, the drawback is that spatial information is lost during pooling. For example, $75\%$ of activations will be discarded if a $2\times2$ pooling window with a stride of 2 is used. To better illustrate, Fig.~\ref{fig:2-a} shows the RFs resulting from first applying $3\times3$ convolution, followed by $2\times2$ max-pooling with stride of 2, followed by $3\times3$ convolution. Inserting a pooling layer increases the size of RF. A $3\times3$ kernel on the higher-level feature map has an $8\times8$ RF on the lower-level feature map. However, the lower level locations that pass on information to the higher levels varies with the input. Successively applying max-pooling layers results in a loss of important spatial information, which we expect will degrade the performance of gaze estimation.
	
	\subsection{Dilated-Convolutions}
	Dilated-convolutional layers preserve spatial resolution while increasing the size of RF without a large increase in the number of parameters. Given an input feature map $(U)$, a kernel of size $N\times M\times K$ (with weights $W$ and bias $b$) and dilation rates  $(r_1, r_2)$, the output feature map $(V)$ of a dilated-convolutional operation can be calculated by
	\begin{equation}
	v(x,y)=\sum_{k=1}^{K}\sum_{m=0}^{M-1}\sum_{n=0}^{N-1}u(x+nr_1,y+mr_2,k)w_{nmk}+b, \label{dilated}
	\end{equation}
	where $(x,y[,k])$ represents the position in the corresponding feature map. Equation~\eqref{dilated} shows that the dilation rates $(r_1, r_2)$ determine the amount by which the size of RF increases. Dilation rates which are larger than one allow the network to enlarge the RF without decreasing the spatial resolution (compared to the use of pooling layers) or increasing the number of parameters.
	
	Fig.~\ref{fig:2-b} shows the RF resulting from a $3\times3$ dilated-convolutional layer with dilation rates $(1, 1)$ followed by a $3\times3$ dilated-convolutional layer with dilation rates $(2, 2)$. Because of the $(2, 2)$ dilation rates, a $3\times3$ kernel applied to the higher-level feature map corresponds to a $7\times7$ RF on the lower-level feature map. Spatial resolution is preserved. The lower level locations feeding into the higher level units are also constant, independent of the input.
	
	Fig.~\ref{fig:2-c} shows the result of successively applying two $3\times3$ convolutional layers while maintaining spatial resolution. The corresponding RF on the lower level map is only $5\times5$. Stacking convolutional layers only increase the size of RF linearly, which makes it hard to cover large regions at higher layers.
	
	\subsection{Dilated-Nets}
	\subsubsection{Multi-Region Dilated-Net.}
\begin{figure}[!t]
		\centering
		\includegraphics[width=12cm]{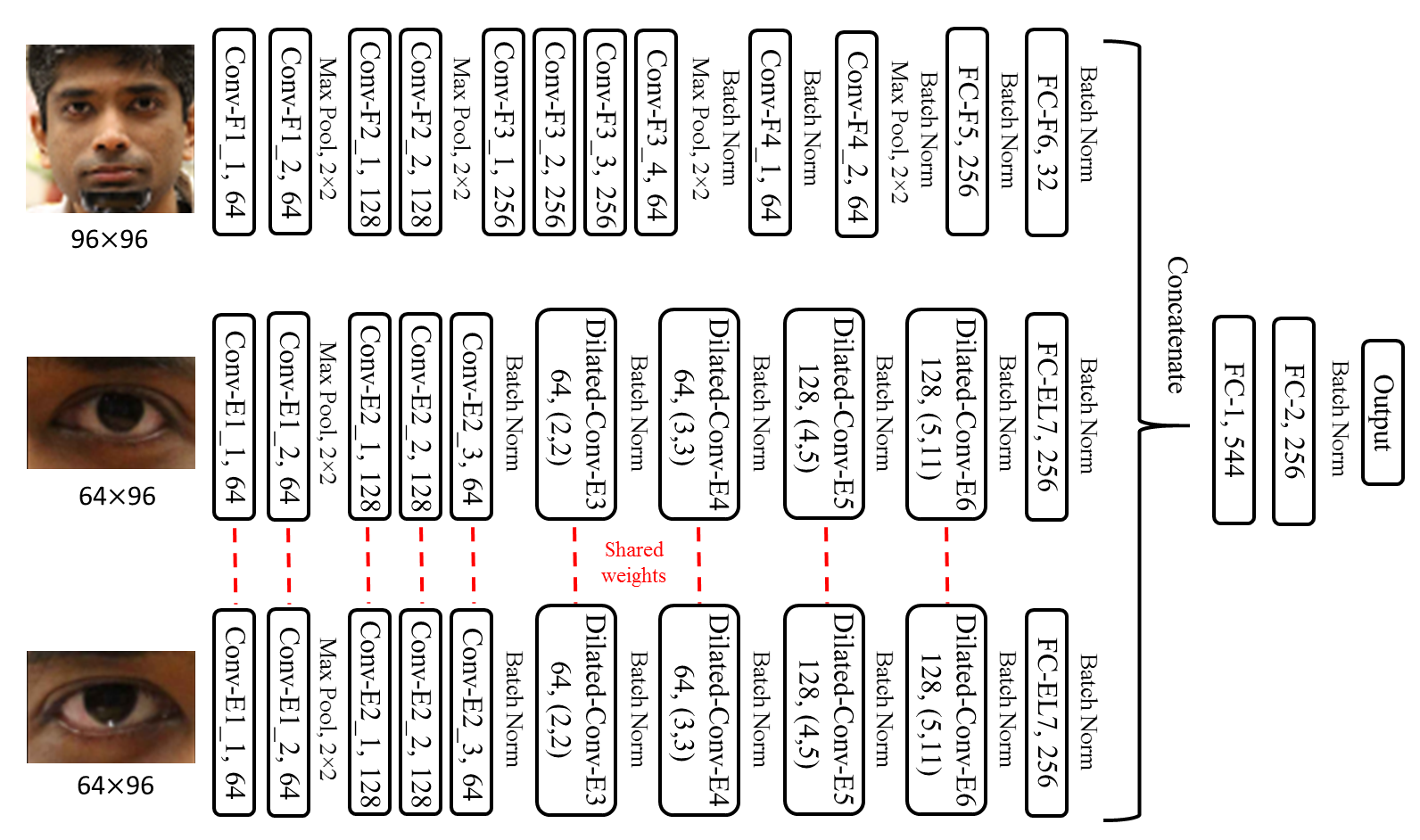}
		\caption{Architecture of the multi-region Dilated-Net. The numbers after convolutional layers (Conv) represent the number of filters. The numbers after dilated-convolutional layers (Dilated-Conv) represent the number of filters and dilation rates. $3\times 3$ filter size is used except that Conv-F3\_4 and Conv-E2\_3 use $1\times 1$. FC represents fully connected layer.}
		\label{fig:3}
	\end{figure}

	Our proposed architecture, which we refer to in our results as \textbf{Dilated-Net (multi)}, is shown in Fig.~\ref{fig:3}. It takes an image of the face and images of both eyes as input and feeds them to a face network and two eye networks, respectively. The general architecture is inspired by iTracker~\cite{krafka2016eye}. However, here the eye networks adopt dilated-convolutional layers to extract high resolution features. 
	
	The face network is a VGG-like network~\cite{simonyan2014very} which consists of four blocks of stacked convolutional layers (Conv) followed by max-pooling layer, as well as two fully connected layers (FC). The weights of the first seven convolutional layers are transferred from the first seven layers of VGG-16 pre-trained on the ImageNet dataset. We insert a $1\times1$ convolutional network (Network in Network~\cite{lin2013network}) after the last transferred layer to reduce the number of channels.
	
	The two eye networks have identical architecture and share the same parameters in all convolutional and dilated-convolutional layers. The network starts with four convolutional layers with a max-pooling layer in the middle, followed by a $1\times1$ convolutional layer, four dilated-convolutional layers (dilated-Conv) and one fully connected layer. The dilation rates of the layers are $(2, 2)$, $(3, 3)$, $(4, 5)$ and $(5, 11)$, respectively. These dilation rate are designed according to the hybrid dilated convolution (HDC) in~\cite{wang2017understanding} so that the RF of each layer covers a square region without any holes in it. The weights of first four convolutional layers are transferred from the first four layers of VGG-16 pre-trained on the ImageNet dataset.
	
	We concatenate the output of FC-F6, FC-EL7 and FC-ER7 to combine features from different input. This is fed to FC-2 and then an output layer. 
	
	We use the Rectified Linear Unit (ReLU) as activation function for all convolutional and fully connected layers. Zero padding is applied to convolutional layers to preserve dimension. No padding is applied to dilated-convolutional layers to reduce the output dimension and computation. We apply the batch renormalization layers (Batch Norm)~\cite{ioffe2017batch} to all layers trained from scratch. Dropout layers are applied to all fully connected layers. 
	\subsubsection{Single-Eye Dilated-Net.}
	To compare with the networks which only use the left eye image and estimated head pose as input, e.g.,~\cite{zhang2017mpiigaze,ranjan2018light}, we shrunk the Dilated-Net (multi) in Fig.~\ref{fig:3} down to only one eye network and concatenated the estimated head pose angles with FC-EL7. In our results, we refer to it as \textbf{Dilated-Net (single)}.
	\subsubsection{CNN without Dilated-Convolutions.}
	To show the improvement achieved by dilated convolutions, we use a deep CNN that has similar architecture but no dilated-convolutions. It replaces the four dilated-convolutional layers by four convolutional layers and three max pooling layers located at the beginning, in the middle and at the end of the four convolutional layers, respectively. The size of final feature maps is the same as the one in Dilated-Nets, and this CNN has the same number of parameters as the corresponding Dilated-Net.
	
\begin{figure}[!t]
		\centering
		\subfloat{		
			\includegraphics[width=3.5cm]{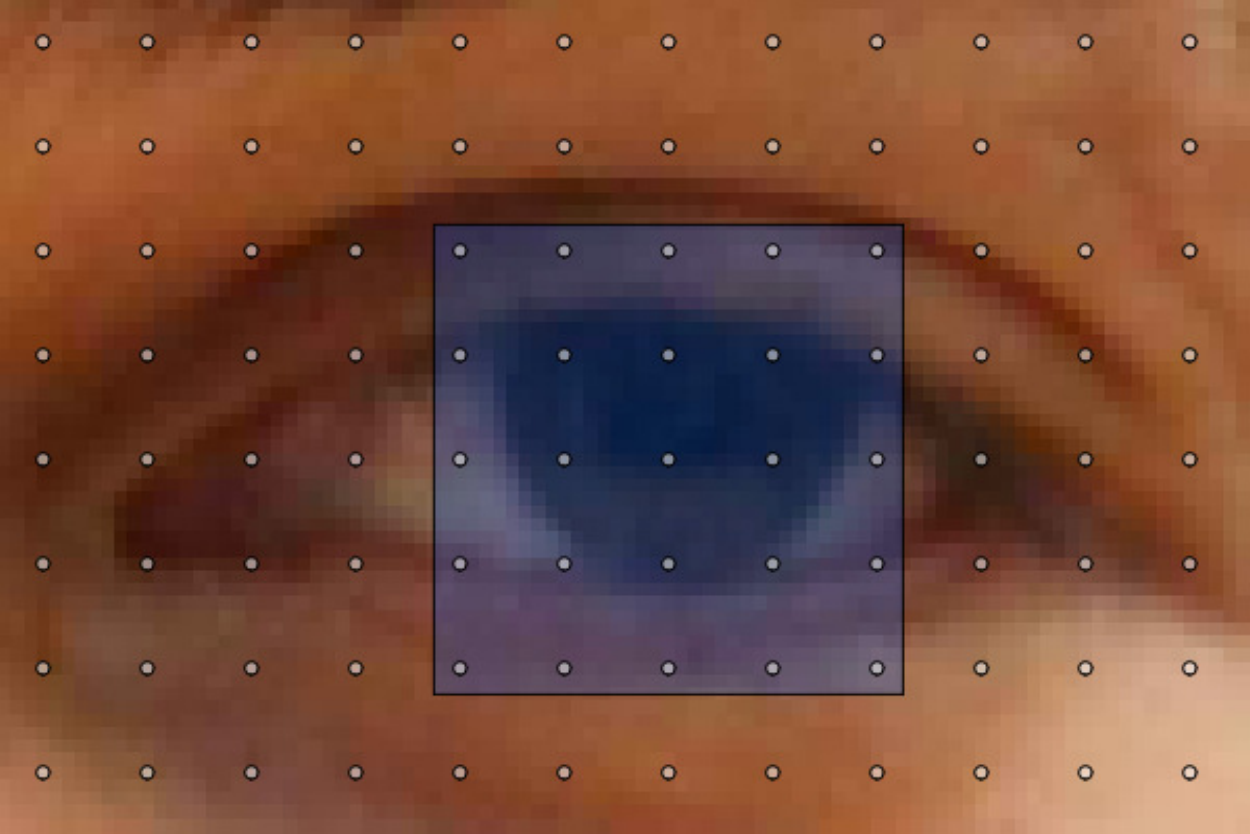}
			\includegraphics[width=3.5cm]{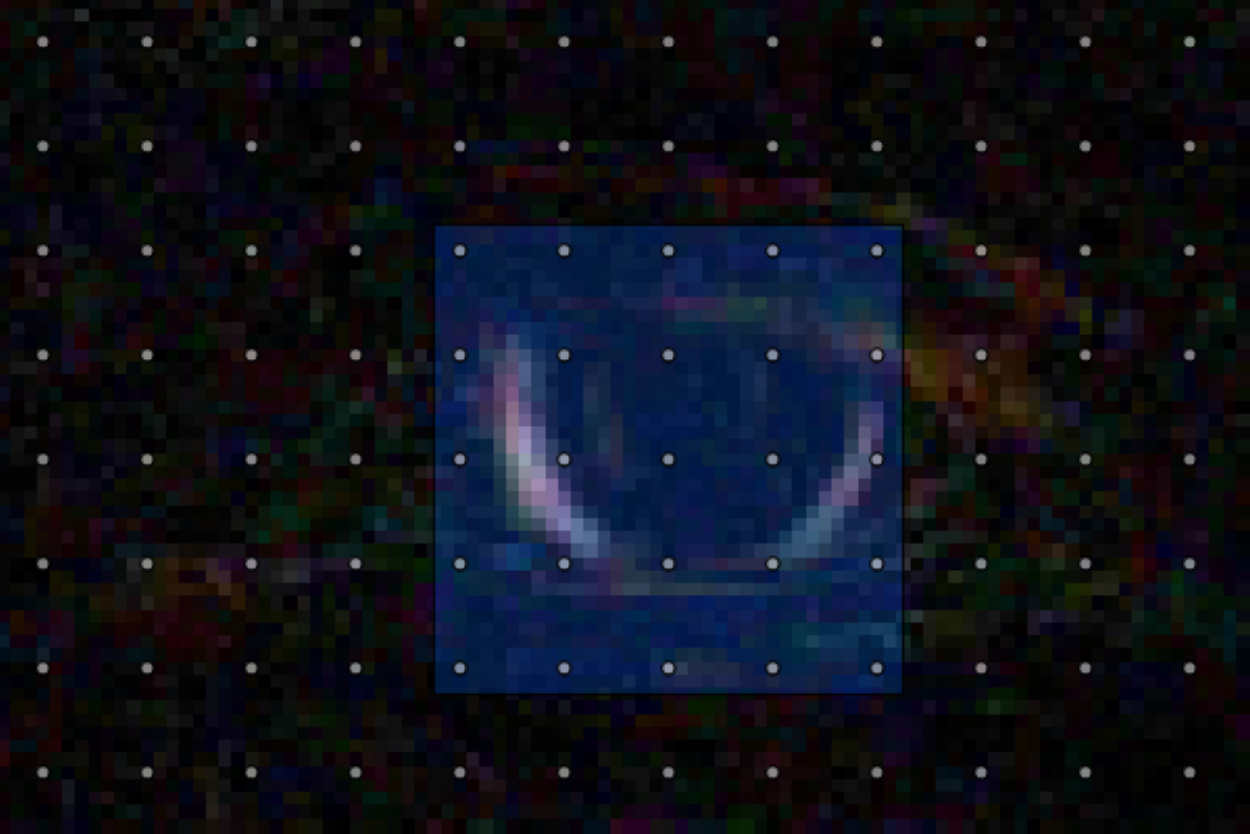}
		}
		\\
		\subfloat{		
			\includegraphics[width=3.5cm]{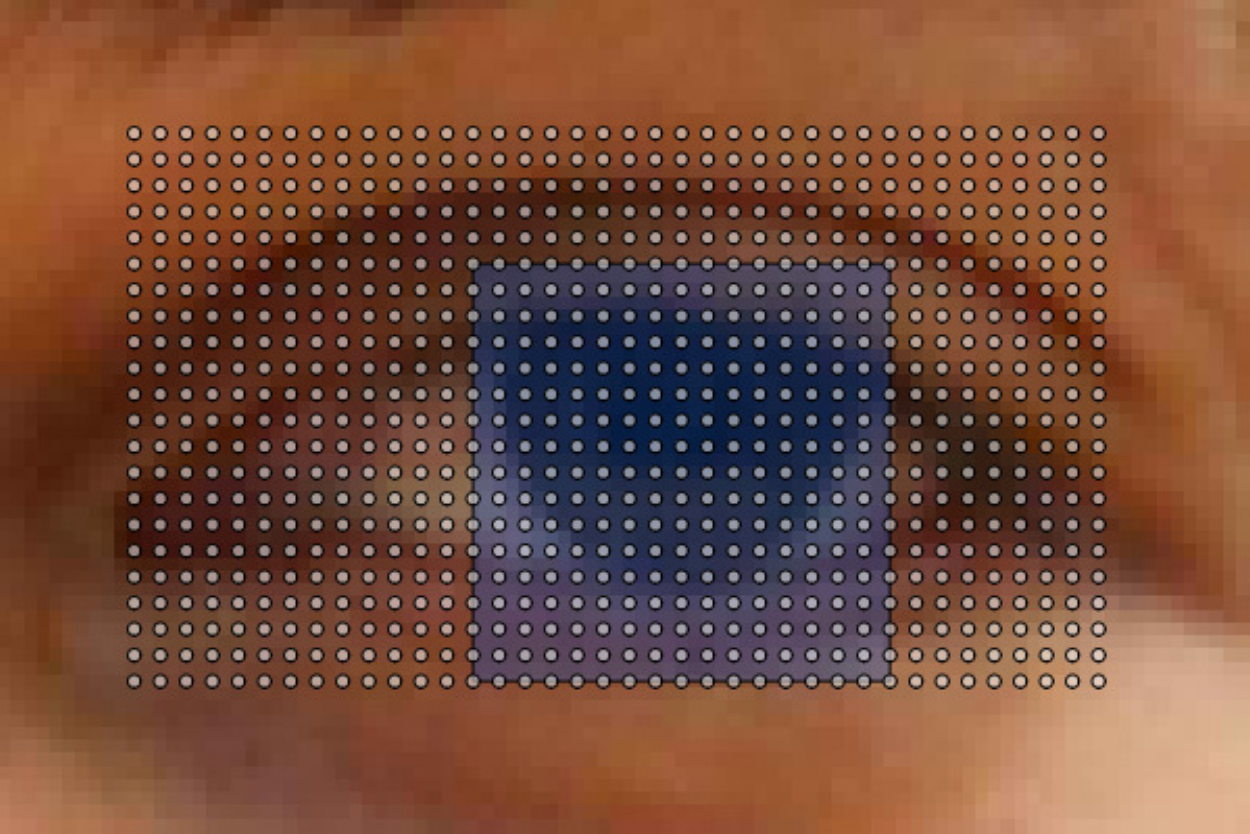}
			\includegraphics[width=3.5cm]{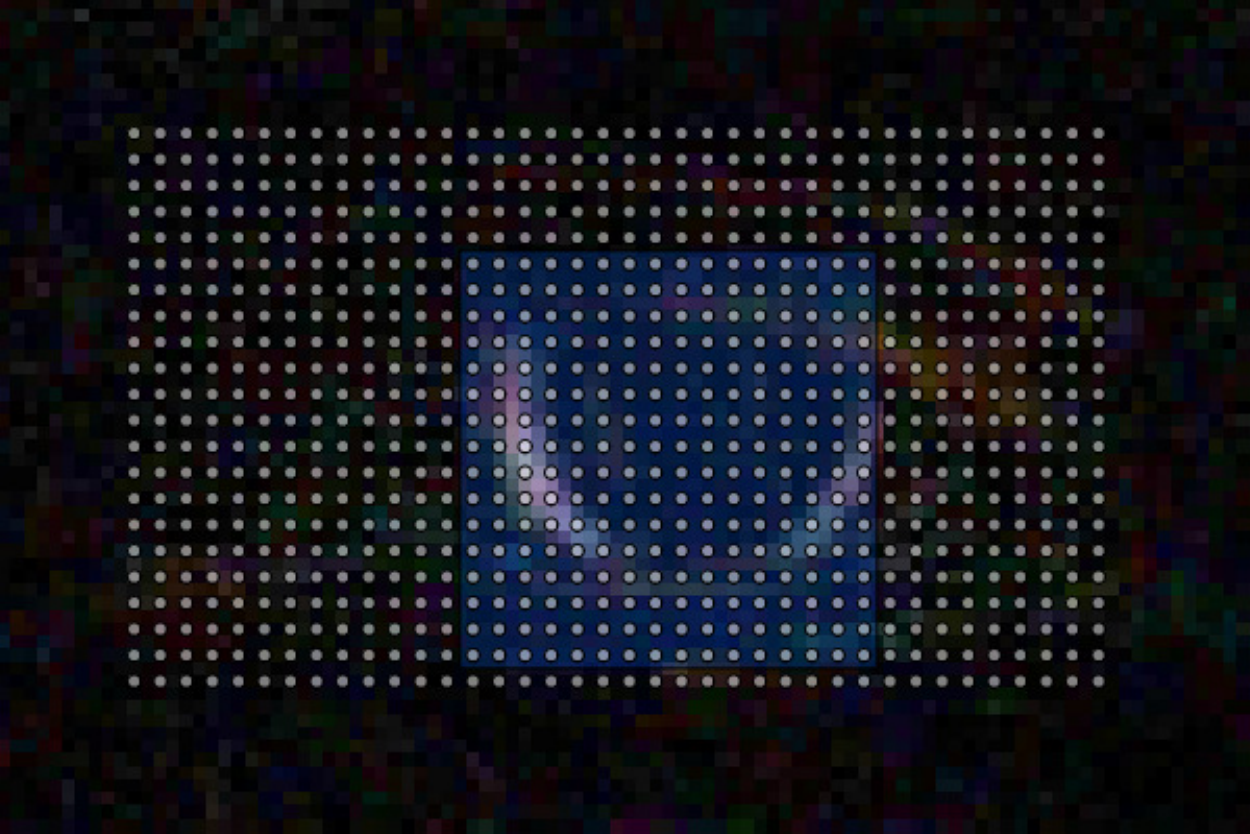}
		}
		
		\caption{Size (blue square) and centers (white dots) of the receptive fields. Top row: The third max-pooling layer in our CNN ($36\times36$ with distance 8). Bottom row: The Dilated-Conv-E4 layer in our Dilated-Net ($34\times34$ with distance 2).}
		\label{fig:4}
	\end{figure}

	One key difference between the CNN and the Dilated-Net is shown in Fig.~\ref{fig:4}, where we show the sizes and centers of RFs of the third max-pooling layer in the CNN and of the Dilated-Conv-E4 layer in Dilated-Net. In the CNN, the progressively use of max-pooling results in larger distance (8 px) between two centers of RFs. For Dilated-Net, the distance between two centers is preserved (2 px). 
	
	\subsection{Preprocessing}
	We apply a two-step preprocessing method. In the first step, We apply the same image normalization method used in~\cite{ranjan2018light,zhang2017mpiigaze,zhang2017s}. This method virtually rotates and translates the camera so that the virtual camera faces the reference point at a fixed distance and cancels out the roll angle of the head. The reference point is set to be the center of the left eye for Dilated-Net (single) and the center of the face for Dilated-Net (multi). The images are normalized by perspective warping, converted to gray scale and histogram-equalized. The estimated head pose angles and the ground truth gaze angles are also normalized. 
	
	In the second step, we obtain the eyes images and face image from the warped images based on facial landmarks. For automatically detected landmarks we use dlib~\cite{king2009dlib}. Then, for each eye image, we use the eye center as the image center and warp the image by an affine transformation so that the eye corners are at fixed positions. For the face image, we fix the position of the center of two eyes and scale the image so that the horizontal distance between the two eye centers is a constant.
	
	\section{Experiments}
	\subsection{Cross-Subject Evaluation}
	We performed cross-subject experiments on the Columbia Gaze dataset~\cite{smith2013gaze} for eye contact detection and the MPIIGaze dataset~\cite{zhang2015appearance} for 3D gaze regression.
	
	\subsubsection{Columbia Gaze Dataset.}
	This dataset was collected for a task to classify whether the subject is looking at the camera (gaze locking). It comprises 5880 full face images of 56 people (24 female, 21 with glasses) taken in a controlled environment. For each person, images were collected for each combination of five horizontal head poses $(0^\circ, \pm15^\circ, \pm30^\circ)$, seven horizontal gaze directions $(0^\circ, \pm5^\circ, \pm10^\circ, \pm15^\circ)$ and three vertical gaze directions $(0^\circ, \pm10^\circ)$. Among the 105 images of each person, five are gaze locking ($0^\circ$ horizontal and $0^\circ$ vertical gaze direction).
	
	\begin{table}[!b]
		\centering
		\renewcommand\arraystretch{1.2}
		\caption{Results on the Columbia Gaze Dataset.}
		\begin{tabular}{c|c|c|c|c|c}
			\Xhline{1pt}
			Name&Method&Input&Training set&PR-AUC&Best F1-score\\
			\hline
			\multirow{2}*{GL~\cite{smith2013gaze}}&\tabincell{c}{PCA+MDA\\+SVM}&Two eyes&Columbia&0.08&0.15\\
			\cline{2-6}
			{}&PCA+SVM&Two eyes&Columbia&0.16&0.25\\
			\hline
			OpenFace~\cite{baltrusaitis2018openface}&Model-based&\tabincell{c}{Two eyes}&SynthesEyes~\cite{wood2015rendering}&0.05&0.10\\
			\hline
			\hline
			\tabincell{c}{CNN (single)\\(ours)}&CNN&\tabincell{c}{Left eye\\+ estimated\\head pose}&Columbia&0.40&0.44\\
			\hline
			\tabincell{c}{Dilated-Net\\(single) (ours)}&Dilated-CNN&\tabincell{c}{Left eye\\+ estimated\\head pose}&Columbia&0.42&0.48\\
			\hline
			\tabincell{c}{CNN (multi)\\(ours)}&CNN&\tabincell{c}{Two eyes\\ + Face}&Columbia&0.48&0.52\\
			\hline
			\tabincell{c}{Dilated-Net\\(multi) (ours)}&Dilated-CNN&\tabincell{c}{Two eyes\\ + Face}&Columbia&\bm{$0.58$}&\bm{$0.62$}\\		
			\Xhline{1pt}		
		\end{tabular}
		\label{table:Columbia-gaze}
	\end{table}
	
	For cross-subject evaluation, we divided 56 subjects into 11 groups, where ten groups contained five subjects and one group contained six. The numbers of male/female subjects with/without glasses were balanced among different groups. We conducted leave-one-group-out cross-validation. In each fold, if a validation set was needed, we randomly select one group from the training set. Since the ratio between negative and positive samples is unbalanced $(20: 1)$, we upsampled the positive examples to balance positive and negative examples by randomly disturbing the facial landmarks.
	
	We used a sigmoid function as output. During training, we used cross-entropy as loss function and stochastic gradient descent with momentum (0.9) to train the network (mini-batch size 64). We used an initial learning rate of 0.01 and multiplied it by 0.5 after every 3000 iterations.
	
	We re-implemented the gaze locking method (GL)~\cite{smith2013gaze} as baseline, which reduces the intensity features by Principal Component Analysis (PCA) and Multiple Discriminant Analysis (MDA), and uses an SVM~\cite{chang2011libsvm} as the final classifier. We applied OpenFace 2.0~\cite{baltrusaitis2018openface} to estimate 3D gaze vectors and calculated the cosine of the angular error from the ground truth.
	\begin{figure}[!t]
		\centering
		\includegraphics[width=12cm]{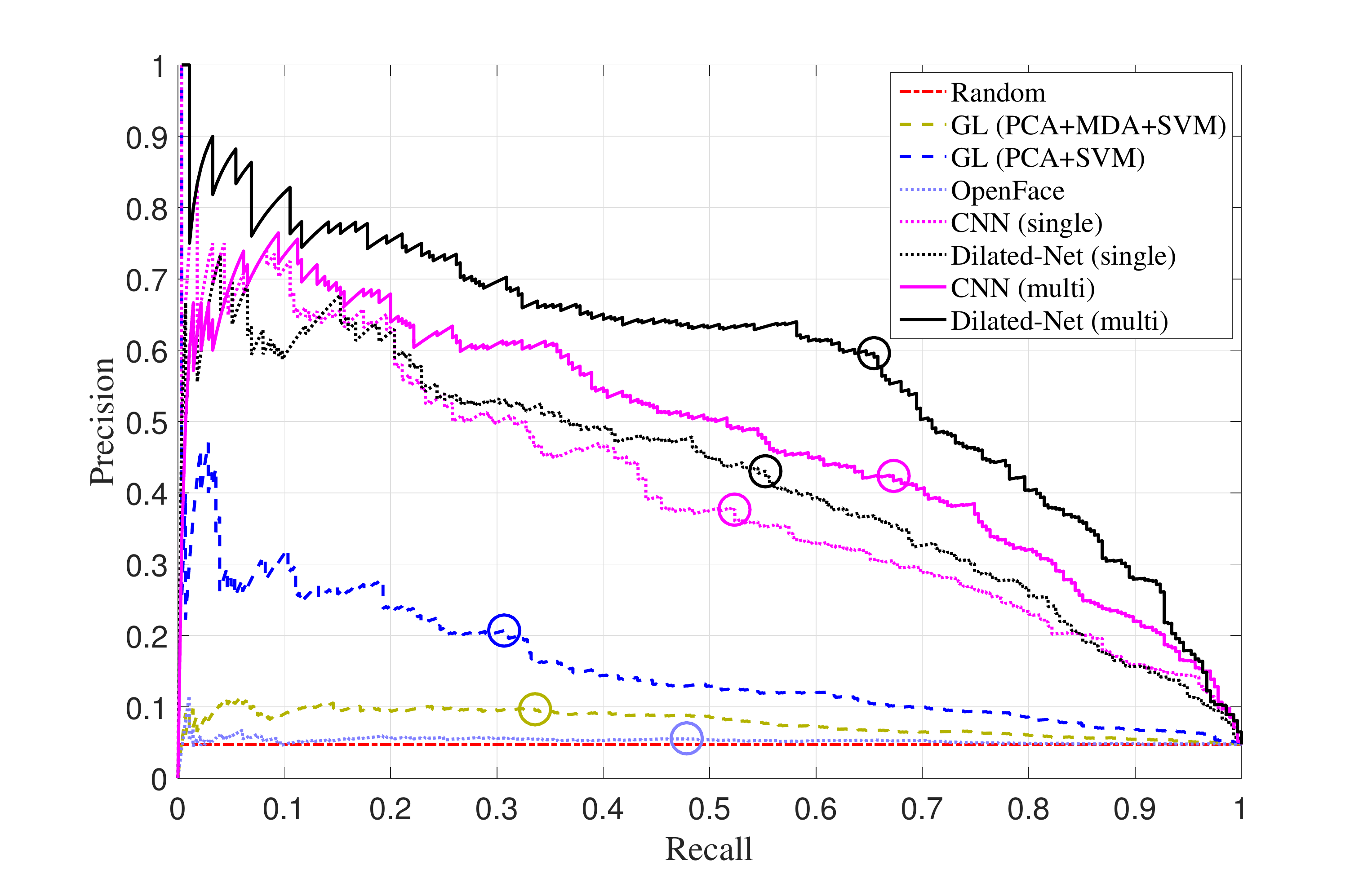}
		\caption{Precision-recall curve of different models on the Columbia Gaze dataset. The circle on each curve indicates the location where the best F1-score is obtained.}
		\label{fig:5}
	\end{figure}
	
	We present the average testing results over 11-folds by the precision-recall (PR) curve in Fig.~\ref{fig:5}. We also report the value of area under curve (PR-AUC) and the best F1-score in Table~\ref{table:Columbia-gaze}. OpenFace performs the worst, mostly because it is a model-based method and trained on another dataset. All deep networks performed better than the two SVM methods, indicating that deep networks have better capacity for appearance-based gaze estimation.
	
	Among the four deep networks, Dilated-Net (multi) performed the best and CNN (single) performed the worst. The degradation of single-eye networks from multi-region networks is because the CNN captures better and more information from the face, which is consistent to~\cite{zhang2017s}. When comparing our Dilated-Net (multi) with the second best model, i.e., CNN (multi), Dilated-Net (multi) outperformed by 0.10 ($20.8\%$) in terms of PR-AUC and by 0.10 ($19.2\%$) in terms of best F1-score. Fig.~\ref{fig:5} shows that Dilated-Net (multi) achieves higher precision for nearly all values of recall, indicating that it generally achieves a better trade-off between precision and recall.
	
	\subsubsection{MPIIGaze Dataset.}
	The MPIIGaze dataset~\cite{zhang2015appearance} was collected for a task to estimate the continuous gaze direction angles. It contains $213,659$ images of 15 subjects (six female, five with glasses). It provides an ``Evaluation Subset'', which contains $3,000$ randomly selected samples for each subject with automatically detected (manually annotated) facial landmarks. As in~\cite{zhang2017mpiigaze,zhang2017s,ranjan2018light}, we trained and tested our Dilated-Net on this ``Evaluation Subset'', which we refer to as MPIIGaze (MPIIGaze+).
	
	We conducted leave-one-subject-out cross-validation. During each fold, we randomly chose the data from three subjects in training set for validation. We trained all our networks with the same validation set in each fold. We used a linear layer to output estimated yaw and pitch gaze angles. The Euclidean distance between the estimated gaze angles and the ground truth angles in normalized space was used as the loss function. We trained all the networks using the Adam optimizer with a mini-batch size 64. An initial learning rate of 0.001 was used. The learning rate was multiplied by 0.1 after every 8000 iterations.
	
	\begin{table}[!b]
		\centering
		\renewcommand\arraystretch{1.3}
		\caption{Mean Angular Errors in Normalized Space Using Eye Center as Origin.}
		\begin{tabular}{c|c|c|c|c|c}
			\Xhline{1pt}
			Name&Architecture&Input&Pre-train&MPIIGaze&MPIIGaze+\\
			\hline
			GazeNet~\cite{zhang2017mpiigaze}&VGG-16&\tabincell{c}{Left eye\\+ estimated\\head pose}&ImageNet&$5.5^\circ$&$5.4^\circ$\\
			\hline
			\multirow{6}*{Branched CNN~\cite{ranjan2018light}}&\multirow{3}*{AlexNet}&\multirow{3}*{\tabincell{c}{Left eye\\+ estimated\\head pose}}&ImageNet&$5.88^\circ$&$5.88^\circ$\\
			\cline{4-6}
			{}&{}&{}&RFC~\cite{su2015render}&$5.56^\circ$&$5.38^\circ$\\
			\cline{4-6}
			{}&{}&{}&\tabincell{c}{RFC\\+ 1M Synth}&$5.38^\circ$&$5.3^\circ$\\
			\cline{2-6}
			{}&\multirow{3}*{\tabincell{c}{AlexNet\\+ branch}}&\multirow{3}*{\tabincell{c}{Left eye\\+ estimated\\head pose}}&ImageNet&$5.77^\circ$&$5.63^\circ$\\
			\cline{4-6}
			{}&{}&{}&RFC&$5.49^\circ$&$5.46^\circ$\\
			\cline{4-6}
			{}&{}&{}&\tabincell{c}{RFC\\+ 1M Synth}&$5.48^\circ$&$5.42^\circ$\\
			\hline
			\hline
			\tabincell{c}{CNN (single)\\ (Ours)}&VGG-16&\tabincell{c}{Left eye\\ + estimated\\head pose}&ImageNet&{$5.45^\circ$}&$5.35^\circ$\\
			\hline
			\tabincell{c}{Dilated-Net(single)\\ (Ours)}&Dilated-CNN&\tabincell{c}{Left eye\\ + estimated\\head pose}&ImageNet&$\bm{5.21^\circ}$&$\bm{5.12^\circ}$\\		
			\Xhline{1pt}		
		\end{tabular}
		\label{table:eye}
	\end{table}
	
	\begin{table}[!b]
		\centering
		\renewcommand\arraystretch{1.2}
		\caption{Mean Angular Errors in Original Space Using Face Center as Origin.}
		\begin{tabular}{c|c|c|c|c}
			\Xhline{1pt}
			Name&Architecture&Input&Pre-train&MPIIGaze+\\
			\hline
			GazeNet~\cite{zhang2015appearance}&AlexNet&\tabincell{c}{Left eye + estimated\\head pose}&ImageNet&$6.7^\circ$\\
			\hline
			iTracker~\cite{krafka2016eye,zhang2017s}&AlexNet&\tabincell{c}{Two eyes\\+ face}&ImageNet&$5.6^\circ$\\
			\hline
			\multirow{2}*{\tabincell{c}{Spatial weights\\ CNN~\cite{zhang2017s}}}&AlexNet&Face&ImageNet&$5.5^\circ$\\
			\cline{2-5}
			{}&\tabincell{c}{AlexNet + sptial\\weights}&Face&ImageNet&$\bm{4.8^\circ}$\\
			\hline
			\hline
			\tabincell{c}{CNN (multi)\\(Ours)}&VGG-16&\tabincell{c}{Two eyes\\+ face}&ImageNet&$5.4^\circ$\\
			\hline
			\tabincell{c}{Dilated-Net (multi)\\(Ours)}&Dilated-CNN&\tabincell{c}{Two eyes\\+ face}&ImageNet&$\bm{4.8^\circ}$\\
			\Xhline{1pt}		
		\end{tabular}
		\label{table:face}
	\end{table}
	
	For the networks that use a single eye image and estimated head pose as input, we compared with the GazeNet~\cite{zhang2017mpiigaze} and the state-of-the-art branched CNN~\cite{ranjan2018light}. For the networks that use images of the full face, we compared with a re-implementation of iTracker~\cite{krafka2016eye} reported in~\cite{zhang2017s} and the state-of-the-art spatial weights CNN~\cite{zhang2017s}. Note that~\cite{ranjan2018light,zhang2017mpiigaze} reported the angular errors in the normalized space and ~\cite{zhang2017s} reported the results in the original space (the camera coordinates).
	
	We present the mean angular errors across 15 subjects in Table~\ref{table:eye} and Table~\ref{table:face}. In Table~\ref{table:eye}, our Dilated-Net (single) achieved the best performance. It achieved $5.21^\circ$ on MPIIGaze and $5.12^\circ$ on MPIIGaze+. It outperformed the second best method, the branched CNN without head-pose-depending-branching, by $0.17^\circ$ ($3.2\%$) on MPIIGaze and by $0.18^\circ$ ($3.4\%$) on MPIIGaze+, even though the branched CNN were pre-trained on more related RFC and 1M synthetic images. The gain was higher if we only considered the networks that were pre-trained on ImageNet. In this case, it outperformed the second best network, CNN (single), by $0.24^\circ$ ($4.4\%$) on MPIIGaze and by $0.23^\circ$ ($4.3\%$) on MPIIGaze+. 
	
	\begin{figure}[!t]
		\centering
		\includegraphics[width=12cm]{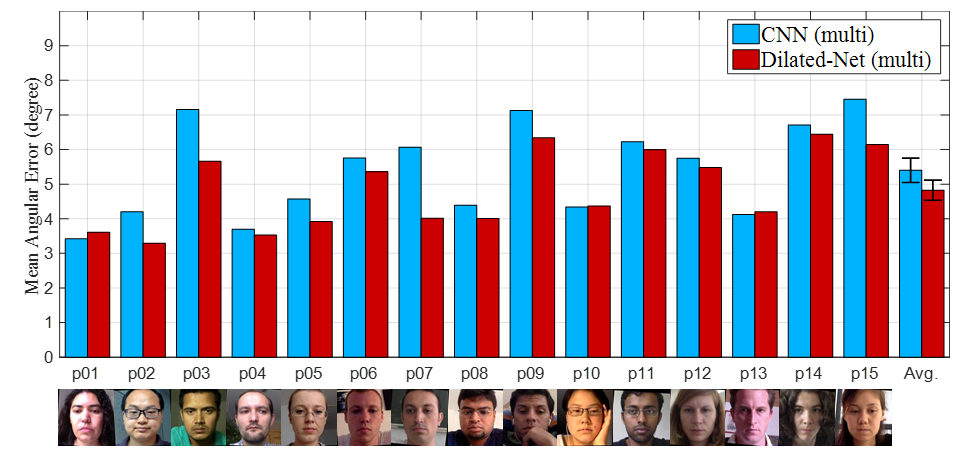}
		\caption{Mean angular error of different subjects on the MPIIGaze dataset in original space. Error bars indicate standard errors computed across subjects.}
		\label{fig:6}
	\end{figure}
	
	Similar results can be observed in Table~\ref{table:face}. When compared to the networks that use similar input (face + two eyes), our proposed Dilated-Net (multi) outperformed iTracker (AlexNet) by $0.8^\circ$ ($14.3\%$) and CNN (multi) by $0.6^\circ$ ($11.1\%$). In Fig.~\ref{fig:6}, we further compare the average error for each subject. Dilated-Net (multi) outperformed CNN (multi) for 12 out of the 15 subjects. Dilated-convolutions improves accuracy for most subjects, despite variations in individual appearance.
	
	Finally, we would like to note that our Dilated-Net (multi) achieved the same results as the state-of-the-art spatial weights CNN. Compared to the spatial weights CNN, our Dilated-Net has several advantages including smaller input size ($96\times224$ v.s. $448\times448$), lower ($64\%$) input resolution, and much smaller number of parameters ($\sim 5$ M v.s. $\sim 196$ M). This suggests that the Dilated-Net might achieve better performance for low resolution images. 
	
	\subsection{Comparing Dilated-CNN with CNN}
\begin{figure}[!t]
		\centering
		\subfloat[]{		
			\includegraphics[width=5.7cm]{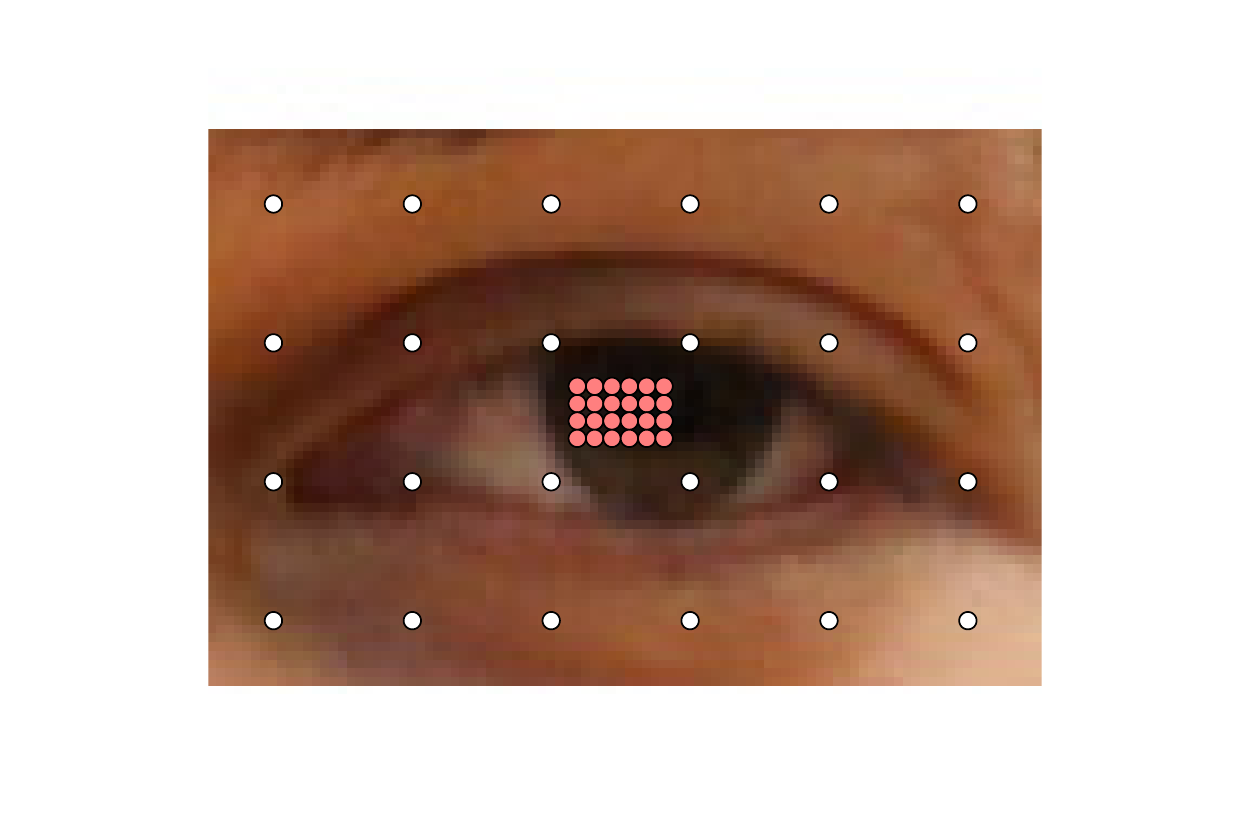}
			\label{fig:7-a}	
		}
		\subfloat[]{		
			\includegraphics[width=6cm]{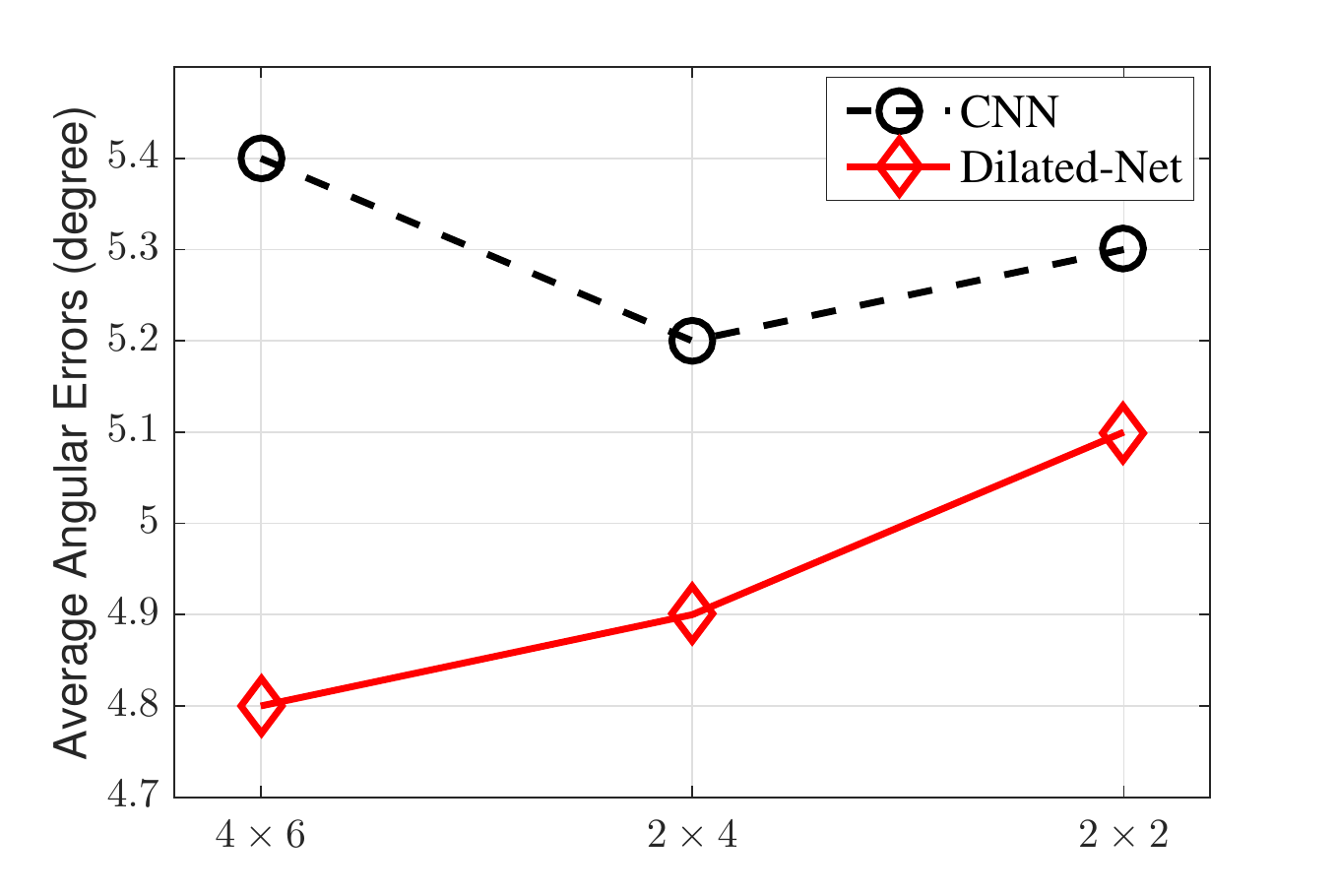}
			\label{fig:7-b}
		}
		\caption{(a) The corresponding center locations of the learned features of the final (dilated-)convolutional layers of the CNN (white) and the Dilated-Net (red); (b) The average angular errors as a function of the remaining features.}
	\end{figure}

	To better understand the differences between Dilated-Net (multi) and CNN (multi), we studied the features learned by the final convolutional layers and evaluated their importance. Both networks have $4\times 6\times 128$ final feature maps. The sizes of the RFs are also similar ($76\times 76$ for the CNN and $70\times 98$ for the Dilated-Net), but they are centered at different locations. The center locations of the CNN units spread over the entire eye image (white dots in Fig.~\ref{fig:7-a}), but the center locations of the Dilated-Net are concentrated at the center (red dots).
	
	We performed an ablation study to determine the contribution of features from different spatial locations, where we only retrained the parameters of the fully connected layers. We left the face network unchanged and evaluated on MPIIGaze+. The average angular errors are presented in Fig.~\ref{fig:7-b} for three cases: (1) using all $4 \times 6$ spatial locations, (2) eliminating the boundary locations and using only the $2 \times 4$ array in the center, and (3) using only the $2\times 2$ array in the center. For the CNN, eliminating the boundary features actually improves performance. This may be due to the removal of person-specific features, enabling better generalization. For the Dilated-Net, we see a degradation in performance as features from different locations are removed. This indicates that despite the significant overlap of the RFs due to the close center spacing, the features at different locations are not redundant. Note that Dilated-Net only using features in $2\times 2$ center region still outperforms the best performing CNN.
	
	We applied t-Distributed Stochastic Neighbor Embedding (t-SNE)~\cite{maaten2008visualizing} to reduce the feature dimension at each location from $128$ to $1$, and used Pearson's r to evaluate the linear correlation between these 1D features and the horizontal gaze angles with fixed head poses and vertical gaze angles. The CNN features were less correlated with gaze. Even restricting attention to the central $2\times 4$ array, correlation coefficients ranged from $0.59$ - $0.82$, whereas for the Dilated-Net, they ranged from $0.74$ - $0.86$ for all 24 locations. 
	
	\subsection{The Effect of Landmarks Precision}
	To evaluate the influence of facial landmark detection, we randomly disturbed the landmarks by $7\%-10\%$  of a $64\times 96$ eye image in each direction. In cross-subject evaluation of MPIIGaze+, performance of the Dilated-Net (multi) degrades by $0.4^\circ$ to $5.2^\circ$ from $4.8^\circ$. CNN (multi) degrades by $0.2^\circ$ to $5.6^\circ$ from $5.4^\circ$. While the Dilated-Net is more sensitive to landmark detection, it is still robust and maintains the performance gains over the CNN.
	
	\subsection{The Use of ResNet}
	To study whether a similar improvement can be obtained using a more advanced architecture, we trained a modified ResNet-50~\cite{He2015} and a Dilated-ResNet on the MPIIGaze+ and tested them on the Columbia Gaze dataset. We changed the stride of the first layer to one and modified the last two residual blocks to dilated-residual blocks. The average angular errors are $6.1^\circ$ (Dilated-ResNet), $6.3^\circ$ (Dilated-VGG),  $6.5^\circ$ (ResNet) and $7.2^\circ$ (VGG). While we can achieve improvement by replacing VGG with the more advanced ResNet, greater improvement is achieved by using dilated-convolutions on VGG. In addition, our results indicate that introducing dilated-convolutions to ResNet further improves performance.

	\section{Conclusion}
	We applied dilated-convolutions in deep neural networks to improve appearance-based gaze estimation. The use of dilated-convolutions allows the networks to extract high level features at high resolution from eye images so that the networks can capture small variability. We conducted cross-subject experiments on the Columbia Gaze and the MPIIGaze datasets. Our results indicated significant gains from the use of dilated-convolutions when compared to CNNs with similar architectures but without dilated-convolutions. These high resolution features improve the accuracy of gaze estimation. Our proposed multi-region Dilated-Net achieved state-of-the-art results on both datasets.
	
	Moving forward, we plan to apply our gaze estimation in real-world settings for human-machine interaction and human-robot interaction. As gaze trajectory is an excellent cue about user intent, the results of gaze tracking or eye contact detection can be used to estimate the user intent. This estimated intent enables the systems to react more naturally and to provide appropriate assistance.
	%
	%
	%

\end{document}